\documentclass[11pt]{article}

\usepackage[preprint]{acl}

\usepackage{times}
\usepackage{latexsym}
\usepackage{amsmath}
\usepackage{enumitem}
\usepackage{tabularx}
\usepackage{multirow}
\usepackage{booktabs}
\usepackage{ulem}
\usepackage{makecell}

\usepackage[T1]{fontenc}

\usepackage[utf8]{inputenc}

\usepackage{microtype}

\usepackage{inconsolata}

\usepackage{graphicx}
\usepackage{subcaption}

\title{Reliable Automated Triage in Spanish Clinical Notes: A Hybrid Framework for Risk-Aware HIV Suspicion Identification
}

\author{Morales-Sánchez, Rodrigo \\
  Dept. of Lenguajes y Sistemas Informáticos, Escuela Técnica Superior de Ingeniería Informática, Universidad Nacional de Educación a Distancia (UNED) / Juan
del Rosal 16, 28040, Madrid, Spain \\
  \texttt{rmorales@lsi.uned.es} \\\And
  Montalvo Soto \\
   Dept. Informática y Estadística, Escuela Técnica Superior de Ingeniería Informática, Universidad Rey Juan Carlos (URJC) / Tulipán
s/n, Móstoles, 28933, Madrid, Spain \\
  \texttt{soto.montalvo@urjc.es} \\\And
  Martínez, Raquel \\
  Dept. of Lenguajes y Sistemas Informáticos, Escuela Técnica Superior de Ingeniería Informática, Universidad Nacional de Educación a Distancia (UNED) / Juan
del Rosal 16, 28040, Madrid, Spain \\
  \texttt{raquel@lsi.uned.es} \\\And}

\author{
    \textbf{Rodrigo Morales-Sánchez\textsuperscript{1,}\thanks{Corresponding author}},
    \textbf{Soto Montalvo\textsuperscript{2}},
    \textbf{Raquel Martínez\textsuperscript{1}},
    \\
    \\
    \normalsize{\textsuperscript{1}Dept. Lenguajes y Sistemas Informáticos, ETSI Informática,}\\
    \normalsize{Universidad Nacional de Educación a Distancia (UNED)}
    \\
    \normalsize{\textsuperscript{2}Dept. Informática y Estadística, ETSI Informática,}\\
    \normalsize{Universidad Rey Juan Carlos (URJC)}
\\
\small{
   \texttt{\textsuperscript{1}\{\textsuperscript{*}\href{mailto:rmorales@lsi.uned.es}{rmorales},\href{mailto:raquel@lsi.uned.es}{raquel}\}@lsi.uned.es, \textsuperscript{2}\href{mailto:soto.montalvo@urjc.es}{soto.montalvo}@urjc.es}
 }\\
}

\begin{document}
\maketitle
\begin{abstract}

Standard clinical Natural Language Processing (NLP) benchmarks often yield inflated metrics by forcing deterministic classification on ambiguous instances, thereby obscuring the clinical risks of overconfident predictions. To bridge this gap, we propose a risk-aware hybrid selective classification framework, evaluated on early Human Immunodeficiency Virus suspicion identification in Spanish clinical notes. Our dual-verification approach explicitly decouples aleatoric uncertainty through Mondrian conformal prediction and epistemic uncertainty using a Multi-Centroid Mahalanobis Distance veto. Empirical evaluations reveal that standard uncertainty metrics and baseline classifiers are structurally insufficient for safe medical triage, suffering severe coverage collapse when forced to operate under strict reliability constraints. In contrast, by demanding that clinical narratives pass both probabilistic and geometric safeguards, the proposed framework successfully isolates a highly trustworthy operational domain. 
The obtained results show that explicit, decoupled uncertainty quantification is essential for translating biomedical NLP into responsible clinical practice.

\end{abstract}

\section{Introduction}




The early diagnosis of Human Immunodeficiency Virus (HIV) remains a critical public health challenge. While automated Natural Language Processing (NLP) systems offer a scalable mechanism to triage Electronic Health Records (EHRs) and allocate diagnostic resources by filtering out clear negative cases \cite{Feller2018UsingAssessment, Morales-Sanchez2024EarlyNotes}, their transition into clinical practice is hindered by rigid evaluation paradigms. 

The prediction task identifies documented clinical suspicion of infection. The supervision signal reflects clinical suspicion evidenced in text rather than confirmed serological status. Since documentation varies in completeness and specificity, suspicion is a subjective, partially observable construct. Consequently, linguistic ambiguity and distributional heterogeneity are intrinsic properties of this problem. When standard deterministic classifiers are forced to make rigid binary decisions on such narratives, they routinely produce overconfident misclassifications, directly conflicting with the safety requirements of clinical medicine \cite{Kelly2019KeyIntelligence, Guo2017OnNetworks}.


To translate biomedical NLP research into responsible practice, AI systems must shift from forced classification to action-oriented selective screening via the classification with rejection paradigm. However, standard Uncertainty Quantification (UQ) methodologies, such as Bayesian approximations or deep ensembles, are structurally insufficient for highly imbalanced clinical text \cite{Abdar2021AChallenges}. By relying solely on predictive entropy, they conflate distinct sources of risk, causing models to either indiscriminately defer true positive cases or unsafely automate complex, out-of-distribution (OOD) narratives.

To address standard uncertainty estimation vulnerabilities, we propose a post-hoc Hybrid Selective Screening Framework that decouples aleatoric and epistemic uncertainties. Our system enforces a dual-verification policy: it utilizes Mondrian Conformal Prediction (MCP) to manage probabilistic ambiguity and a Multi-Centroid Mahalanobis Distance (MCMD) veto within a spectrally normalized latent space to filter epistemic anomalies. This intersectional safeguard translates risky forced-binary predictions into safe, trinary triage recommendations.


The main contributions of this work are:
\begin{itemize}
    \item \textbf{Hybrid Selective Screening Framework:} We develop a novel, dual-veto deferral system that explicitly combines conformal prediction for aleatoric ambiguity with a geometrically constrained feature space for epistemic anomaly detection.  We show that the strict intersection of both safeguards is essential for safe clinical automation.
    \item \textbf{Subjective Construct Modeling in Spanish EHRs:} We formalize the extraction of early HIV clinical suspicion as a partially observable inference task, addressing the linguistic and epistemological challenges of clinical NLP in a resource-constrained, non-English environment. 
    \item \textbf{Rigorous Clinical Validation:} Utilizing asymmetric, risk-aware clinical metrics (e.g., Custom Risk-Kappa, True Positive Deferral Rate), we empirically expose the operational failure of standard UQ algorithms on clinical data. We prove that our proposed hybrid policy significantly outperforms standard baseline UQ methods, maintaining robust calibration without sacrificing triage efficiency.
\end{itemize}

\section{Related Work}

\textbf{Early HIV Prediction via NLP} \quad 
Historically, automated HIV risk assessment relied on structured EHR fields, which consistently underperform as stigmatized behavioral risks are rarely captured in standard coding \cite{Feller2018UsingAssessment}. Consequently, research has pivoted toward NLP to extract nuanced indicators from unstructured narratives using contextual word embeddings \cite{Morales-Sanchez2024EarlyNotes,Sah2025RoleOutcomes}. As these predictive architectures mature, there is a consensus that deterministic models require explicit reliability frameworks for safe clinical adoption \cite{Ngema2026ExplainableStudy}.


\textbf{Reliability and Selective Classification} \quad 
The deployment of AI in high-stakes public health settings requires rigorous UQ \cite{Abdar2021AChallenges}. 
Recent work has applied selective classification to clinical text, demonstrating that deep learning classifiers can achieve strict safety targets by abstaining from unreliable samples without the prohibitive computational burden of model retraining \cite{Peluso2024DeepClassification, Garcia2026SelectiveSetting}.
However, standard UQ methodologies struggle with highly imbalanced clinical text because their reliance on predictive entropy conflates aleatoric and epistemic uncertainty \cite{Ziletti2026}. In medical triage, this conflation is operationally fatal, leading to indiscriminate deferrals or unsafe automation \cite{Toure2026}. Because safe clinical deployment demands models that explicitly know when to abstain \cite{Machcha2025}, there is a pressing need for hybrid architectures that decouple and independently bound these distinct predictive risks.

\section{Material and Methods}
\subsection{Hybrid Selective Screening and Deferral Policy}
We implement a post-hoc Dual-Verification Selective Classification that requires a text to pass two distinct uncertainty checks before permitting an automated triage decision.

\begin{figure*}[t]
    \centering
    \includegraphics[width=\textwidth]{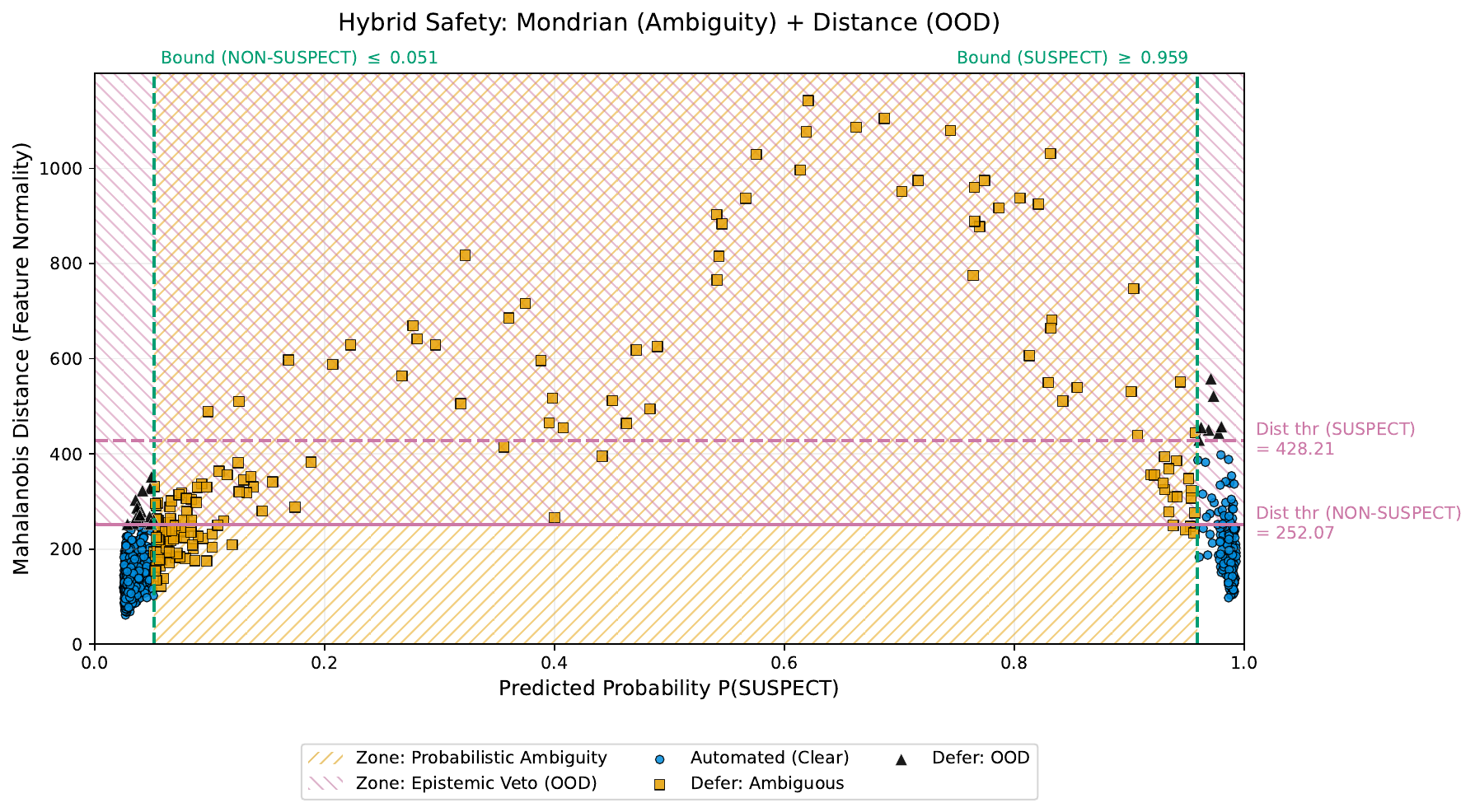}
    \caption{Hybrid Selective Screening Policy in the spectrally normalized latent space. Soft teal vertical dashed lines denote MCP boundaries for probabilistic ambiguity, mapped to the diagonally hatched zone. Reddish-purple horizontal lines indicate Mahalanobis distance thresholds ($\tau_{dist}$) for the epistemic veto, mapped to the cross-hatched zone. Automated `Clear' cases are shown as medium blue circles, probabilistically ambiguous deferrals as golden orange squares, and epistemic OOD deferrals as solid black triangles.}
    \label{fig:example}
\end{figure*}

\textbf{Aleatoric Control via Mondrian Conformal Prediction (MCP)} \quad To manage aleatoric uncertainty arising from inherent ambiguity in the clinical narratives, we employ MCP \citep{Bostrom2021MondrianDistributions}. 
First, we apply Temperature Scaling \citep{Guo2017OnNetworks} to the validation logits to minimize the Negative Log Likelihood (NLL) and mitigate overconfidence. Subsequently, 
MCP generates a prediction set $\Gamma^{\alpha}(x)\subseteq\{0,1\}$ containing the true label with probability $1-\alpha$. We use the set's cardinality to indicate ambiguity: sets containing multiple classes ($|\Gamma^{\alpha}(x)|>1$) or zero classes ($|\Gamma^{\alpha}(x)|=0$) signify insufficient or contradictory evidence at the target risk tolerance $\alpha$, necessitating a deferral.

\textbf{Epistemic Control via Geometric Veto} \quad Deep neural networks frequently suffer from the phenomenon of assigning high softmax probabilities to anomalous, OOD inputs. To mitigate this, we integrate a geometric veto mechanism that operates directly within the $L_2$-normalized latent feature space. Because clinical ``suspicion'' encompasses highly heterogeneous patient presentations—ranging from varying risk behaviors to sparse opportunistic infections—assuming a unimodal distribution for the feature space leads to artificially inflated variances. 
To accurately accommodate this phenotypic diversity without risking the severe overparameterization of local density estimation in a high-dimensional space, we model the clinical narratives using a MCMD framework (effectively a tied-covariance Gaussian Mixture).

For each class $c$, the optimal number of local centroids, $K_c$, is dynamically determined via k-means clustering by identifying the smallest $K$ where the relative inertia gain falls below a predetermined 0.05 threshold, subject to a minimum cluster size constraint. 
Rather than estimating independent, unstable covariance matrices for every local cluster, we assume a globally shared noise structure. By pooling the residuals of all samples to their assigned centroids, a single, robust global precision matrix $\Sigma^{-1}$ is fitted across all classes using Oracle Approximating Shrinkage (OAS).

During inference, the epistemic uncertainty of a sample $x$ is quantified as the minimum Mahalanobis distance between its embedding $f(x)$ and any centroid $\mu_{\hat{y},k}$ associated with its predicted class~$\hat{y}$:
\begin{equation*}
d_{M}(x,\hat{y}) = \min_{k\in\{1...K_{\hat{y}}\}} \sqrt{r_{\hat{y},k}^{\top}\Sigma^{-1}r_{\hat{y},k}}
\end{equation*}
where $r_{\hat{y},k} = f(x) - \mu_{\hat{y},k}$ represents the residual vector between the sample embedding and the $k$-th centroid.

To account for the asymmetric variance between routine (Class 0) and highly heterogeneous suspected cases (Class 1), we establish class-specific distance thresholds $\tau_{dist}^{(\hat{y})}$, calibrated to the 99th percentile of intra-class distances from correctly classified validation samples. Inputs exceeding this boundary are flagged as geometrically anomalous. This mechanism effectively vetoes overconfident OOD predictions without disproportionately deferring the diverse minority class.

These specific thresholds were not chosen arbitrarily; a post-hoc grid search over the validation folds confirmed that the 0.05 inertia and 99th percentile boundaries empirically maximized the Custom Risk-Kappa, optimally balancing OOD anomaly detection with minority class variance (see Appendix \ref{sec:hyperparameter_analysis} for the full sensitivity analysis).

\textbf{Global Decision Logic} \quad The final deferral policy $D(x)$ enforces a strict intersection of both safeguards:
\begin{equation*}
D(x) = 
\begin{cases} 
\hat{y} & \text{if } |\Gamma^\alpha(x)| = 1 \\
        & \text{AND } d_M(x, \hat{y}) \le \tau_{dist}^{(\hat{y})} \\[1ex]
\perp   & \text{otherwise} 
\end{cases}
\end{equation*}


A complete visual representation of this dual-verification logic 
is provided in Figure \ref{fig:example}.

\subsection{Data}
\begin{table*}[t]
\centering
\resizebox{0.8\linewidth}{!}{
\begin{tabular}{@{}l|c|ccc@{}}
\toprule
\textbf{Metric} & \textbf{Total} & \textbf{Train (Avg.)} & \textbf{Validation (Avg.)} & \textbf{Test (Held-out)} \\
\midrule
\textbf{Total Patients} & 13,642 & 11,046 & 1,223 & 1,373 \\[0.5ex]
\quad \textbf{Total Suspects} & \textbf{2,072} & \textbf{1,671} & \textbf{183} & \textbf{218} \\
\quad \textit{HIV+ / HIV-} & \textit{172 / 1,900} & \textit{136 / 1,535} & \textit{14 / 169} & \textit{22 / 196} \\[0.5ex]
\quad Non-Suspects & 11,570 & 9,375 & 1,040 & 1,155 \\[0.5ex]
Class Ratio (Suspect : Non-Suspect) & 1:5.5 & 1:5.6 & 1:5.7 & 1:5.3 \\
\midrule
\textbf{Total Clinical Notes} & 63,802 & 51,642 & 5,710 & 6,450 \\
Avg. Notes / Patient & - & 4.68 & 4.67 & 4.70 \\
Tokens / Patient (Mean / Max) & - & 409 / 8,621 & 407 / 8,033 & 427 / 5,255 \\
Avg. Tokens / Note (Susp. / Non-Susp.) & - & $\sim$126 / $\sim$79 & $\sim$118 / $\sim$79 & $\sim$127 / $\sim$81 \\
\bottomrule
\end{tabular}
}
\caption{Dataset total and partitioning with key characteristics. Training and validation figures represent the average across the 10-fold cross-validation splits.}
\label{tab:dataset_stats}
\end{table*}

The study utilizes a proprietary cohort of Spanish EHRs sourced from Hospital Universitario Fundación Alcorcón (HUFA, Madrid), comprising real-world clinical notes. The total collection comprises 13,642 patients with notes between 1998 and 2023 from different medical specialities. All data was anonymized at the source by the hospital in strict compliance with the 
GDPR and Spanish LOPDGDD regulations by replacing 29 different entity types (including age and gender) with placeholder tags.

Rather than relying on manual annotation, ground truth was established empirically based on documented physician behavior: the suspect cohort comprises patients who had a requested HIV serology (both positive and negative outcomes). EuroTEST \cite{EuroTEST2016HIVSettings} indicator conditions were then applied as a secondary filter to isolate genuine clinical suspicion, excluding cases where serology was requested as a routine administrative procedure. Because the threshold for ordering a test varies among physicians, this supervisory signal inherently captures the subjective, partially observable nature of real-world clinical suspicion, further motivating our selective screening paradigm. However, because clinical documentation practices and physician thresholds for ordering serology vary substantially across institutions, it is important to note that the geometric thresholds calibrated on this single-center cohort would likely require recalibration prior to external deployment. 

To prevent explicit diagnostic leakage and force the model to learn the implicit digital phenotype of early HIV risk, the corpus was aggressively sanitized by removing all explicit HIV acronyms and direct testing recommendations. The non-suspect cohort included patients with no history of such requests. The data was partitioned using 10-fold cross-validation alongside a held-out test set~\allowbreak(Table~\ref{tab:dataset_stats}).

\subsection{Model Architecture}

\subsubsection{Attention-MIL Architecture}
To handle EHR structural variability (Table \ref{tab:dataset_stats}) without discarding longitudinal context, we frame HIV suspicion detection as a Multiple Instance Learning (MIL) problem, adapting \citet{Ilse2018Attention-basedLearning}. 

In this approach, each patient's record is treated as a `bag' of overlapping text sequences. Documents are partitioned using a sliding window strategy (up to 64 chunks of 384 tokens) 
and independently encoded using the Spanish biomedical RoBERTa model \texttt{PlanTL-GOB-ES/bsc-bio-ehr-es}\footnote{\href{https://huggingface.co/PlanTL-GOB-ES/bsc-bio-ehr-es}{PlanTL-GOB-ES/bsc-bio-ehr-es} - Hugginface. Accessed: [2026-02-26]} \cite{Carrino2022PretrainedSpanish}. To aggregate these chunks into a unified document representation $\mathbf{z}_{doc}$, we employ a gated attention pooling mechanism. The network assigns a learned scalar weight to each chunk via a gated combination of hyperbolic tangent and sigmoid activations, ensuring that highly suspicious text snippets dominate the final document-level embedding before it is passed to the UQ evaluation.

%


\subsubsection{Training for Reliable Uncertainty}
\label{sec:models}
To address the class imbalance while maintaining calibration, we contrast two architectures:

\textbf{Standard MIL:} 
Optimized using a Label-Smoothed Focal Loss \cite{Lin2018FocalDetection} to down-weight easy negatives, coupled with Regularized Dropout (R-Drop) \cite{Liang2021R-Drop:Networks} to enforce KL-divergence consistency between forward passes, thereby smoothing the decision boundary.

\textbf{MD-SN MIL:} While Focal Loss improves raw classification, it can distort the underlying probabilities required for aleatoric UQ. Therefore, we utilize the MD-SN (Mahalanobis Distance + Spectral Normalization) architecture proposed by \cite{Vazhentsev2022UncertaintyDetection}, coupled with Logit Adjustment \citep{Menon2021Long-tailAdjustment}. 
This technique adds logarithmic class priors directly to the raw logits, inherently pushing the decision boundary away from the minority class without sacrificing probabilistic integrity.  
To enable the epistemic geometric veto, SN is applied to the dense layers of the MIL head, bounding the Lipschitz constant to ensure a bi-Lipschitz smooth feature space.


\subsection{Uncertainty Estimation Backends}
To demonstrate that our selective screening framework is adaptable and to evaluate the trade-off between computational cost and empirical safety, we extract predictive uncertainty estimates using three established methodologies:

\paragraph{Standard UQ Backends} 
We use Monte Carlo (MC) Dropout \cite{Gal2016DropoutLearning} as a baseline Bayesian approximation, and Cross-Validation (CV) Deep Ensembles \cite{Lakshminarayanan2017SimpleEnsembles} as our empirical gold standard. The first executes $T$ stochastic forward passes during inference to compute predictive entropy as an epistemic uncertainty threshold. The latter aggregates predictions from $K$ models trained on distinct CV folds. Crucially, we utilize an Out-Of-Fold (OOF) calibration strategy to fit deferral thresholds strictly on unseen validation folds, preventing optimistic bias.

\paragraph{Mahalanobis Distance with Spectral Normalization (MD-SN)}
To circumvent the computational overhead of ($T$ times) sampling or (O(K) latency and storage costs of 10 models) ensembling, we evaluate deterministic, single-pass UQ using the MD-SN architecture \cite{Vazhentsev2022UncertaintyDetection}. Unlike entropy-based methods that rely on softmax dispersion, MD-SN measures uncertainty geometrically. By calculating the Mahalanobis distance between a test embedding and the fitted class centroids in a bi-Lipschitz smooth feature space, texts that are semantically atypical or OOD yield proportionally large geometric distances.

\subsection{Evaluation Framework and Clinical Metrics}
Standard NLP evaluation frameworks often fail to capture the asymmetric risks inherent in medical triage. For example, standard metrics like AUROC can be highly misleading on datasets with severe class imbalance, while the standard $F_1$-score treats false positives and false negatives equally. To accurately assess both the discriminative power and the safety of our hybrid selective screening policy, we employ a specialized suite of clinical and probabilistic metrics.

Therefore, we evaluate: (1) Discriminative Safety via $F_{2}$-score, penalizing false negatives (misclassifying a suspect patient as non-suspect) more heavily than false positives; (2) Probabilistic Reliability via Expected Calibration Error (ECE) \cite{Guo2017OnNetworks};  (3) Triage Efficiency via Coverage (fraction of automated records) and True Positive Deferral Rate (TPDR), defined as the proportion of actual suspect cases referred to a physician; and (4) Uncertainty Ranking via the Area Under the Risk-Coverage curve (AURC).

Finally, to evaluate the final trinary triage decision (Clear Negative, Defer, Clear Positive) against the binary ground truth, we compute a (5) Custom Risk-Kappa score. This adaptation of Cohen's Weighted Kappa applies a strictly asymmetric penalty matrix $W$ derived from the established preventative medicine paradigm for infectious diseases. In early HIV detection, the epidemiological cost of a missed diagnosis—delaying ART and risking transmission—far outweighs an unnecessary routine test. To formalize this, the maximum penalty of 1.0 is strictly reserved for the catastrophic failure of automating a False Negative (FN) ($W_{FN}=1.0$). Conversely, False Positive automations ($W_{FP}=0.5$) and the deferral of True Positive cases ($W_{def\_pos}=0.5$) incur moderate penalties, reflecting the lower clinical cost of resource utilization and the failure to automate complex cases. Safely deferring a True Negative carries the lowest penalty ($W_{def\_neg}=0.25$). 

\begin{equation*}
    W = \begin{bmatrix}
    0.0 & 0.5 & 1.0 \\
    0.5 & 0.25 & 0.0 
    \end{bmatrix}
\end{equation*}

A complete sensitivity analysis of these penalty weights across diverse simulated health-economic realities is provided in Appendix \ref{sec:kappa_ablation}.

\section{Results and Discussion}
To evaluate the clinical viability of the proposed selective screening framework, we benchmarked 
the two architectures explained in Section \ref{sec:models}. Also, to isolate the value of our specialized architecture, we introduce a standard transformer baseline using only the base encoder (\texttt{PlanTL-GOB-ES/bsc-bio-ehr-es}).

\begin{table*}[t]
\centering
\resizebox{\textwidth}{!}{
\begin{tabular}{llcccccccc}
\toprule
\multirow{2}{*}{\textbf{Architecture}} & \multirow{2}{*}{\textbf{UQ Algorithm}} & \multicolumn{2}{c}{\textbf{Discriminative ($F_2 \uparrow$)}} & \multicolumn{2}{c}{\textbf{Calibration (ECE $\downarrow$)}} & \multicolumn{4}{c}{\textbf{Triage Efficiency \& Safety}} \\
\cmidrule(lr){3-4} \cmidrule(lr){5-6} \cmidrule(lr){7-10}
& & \textbf{Binary} & \textbf{Clear} & \textbf{Binary} & \textbf{Clear} & \textbf{Coverage $\uparrow$} & \textbf{TPDR $\downarrow$} & \textbf{AURC $\downarrow$} & \textbf{Risk-Kappa $\uparrow$} \\
\midrule
\textbf{Enc. Baseline}
& CV Ensemble & 0.769 & 0.973 & 0.077 & 0.031 & 47.8\% & 50.5\% & 0.026 & 0.377 \\
\midrule
\multirow{2}{*}{\textbf{Standard MIL}} 
& MC Dropout & 0.835 & 0.975 & 0.033 & 0.011 & 68.8\% & 34.2\% & 0.007 & 0.579 \\
& CV Ensemble & 0.823 & 0.979 & \textbf{0.018} & \textbf{0.009} & 63.5\% & \textbf{29.6\%} & 0.007 & 0.546 \\
\midrule
\multirow{3}{*}{\textbf{MD-SN MIL}}
& MC Dropout & \textbf{0.839} & 0.973 & 0.044 & 0.031 & 70.2\% & 30.6\% & 0.011 & \textbf{0.602} \\
& MD-SN & 0.813 & 0.966 & 0.028 & 0.026 & \textbf{70.7\%} & 31.9\% & 0.012 & 0.601 \\
& CV Ensemble & 0.821 & \textbf{0.982} & 0.022 & 0.021 & 67.7\% & 33.2\% & \textbf{0.006} & 0.576 \\
\bottomrule
\end{tabular}
}
\caption{Discriminative performance, calibration, and triage efficiency under strict safety constraints ($\alpha=0.01$). Metrics evaluate the forced deterministic output (Binary) and the automated subset after deferral (Clear). \textbf{Note:} Full results, including 95\% bootstrap Confidence Intervals, are detailed in Appendix \ref{sec:results}.}
\label{tab:strict_safety}
\end{table*}

The architectures were systematically evaluated across multiple UQ algorithms and varying risk tolerances ($\alpha$). 
To rigorously validate statistical stability, the exhaustive evaluation, including 95\% Confidence Intervals (CI) calculated via empirical bootstrap resampling ($n=1,000$ iterations), is provided in Appendix \ref{sec:results}.

\subsection{Forced Binary Classification}
Table \ref{tab:strict_safety} presents a head-to-head comparison under strict safety constraints ($\alpha=0.01$). As detailed in the ``Binary'' columns
, evaluating the cohort under a forced-classification paradigm reveals the exact vulnerability of standard NLP benchmarks. For instance, utilizing MC Dropout on the MD-SN architecture yields the highest absolute Binary $F_2$ score (0.839), yet this masks severely degraded calibration (ECE: 0.044) compared to the MD-SN ensemble (0.022).

A clear architectural progression is visible across the 10-model ensembles. 
Although the MIL architectures demonstrate a clear representational superiority over the Naive Encoder baseline ($\sim$0.05 $F_2$ improvement), their absolute global $F_2$ ($\sim$0.82) remains fundamentally unsafe for autonomous deployment. This underscores the danger of compelling a system to guess on ambiguous or OOD narratives, predictably resulting in overconfident misclassifications.

\subsection{Selective Screening}


Applying the Dual Veto policy at a strict safety tolerance ($\alpha=0.01$) exposes the operational flaws of standard baselines. 
Hampered by poor calibration and sub-optimal uncertainty ranking, the ensembled Naive Encoder suffers a severe coverage collapse (47.8\%) and a clinically unviable TPDR of 50.5\%. In contrast, the ensembled MIL architectures successfully isolate highly reliable predictive domains. 
MD-SN MIL yields the best uncertainty ranking (AURC: 0.006 vs. 0.007 for Standard MIL) and 0.982 Clear $F_2$ at 67.7\% coverage , effectively vetoing overconfident misclassifications.


Comparing UQ backends highlights a critical operational trade-off. While CV-Ensembles establish the empirical safety upper bound, they incur a prohibitive 10x computational overhead. Conversely, MC Dropout bypasses this footprint but struggles with probabilistic calibration in this highly imbalanced domain. Bridging this gap, the deterministic MD-SN UQ algorithm achieves a highly competitive Clear $F_2$ of 0.966 with superior calibration in a single pass. This efficiently isolates the safe operating domain without the stochastic latency of MC Dropout or the memory burden of ensembles, offering a viable alternative for resource-constrained deployments.


%

\subsection{The Necessity of the Dual Veto}
\begin{table}[t]
\centering
\resizebox{\columnwidth}{!}{
\small
\setlength{\tabcolsep}{6pt}
\begin{tabular}{lccc}
\toprule
\textbf{Deferral Policy} & \textbf{Coverage $\uparrow$} & \textbf{Clear F2 $\uparrow$} & \textbf{Risk-Kappa $\uparrow$} \\
\midrule
\multirow{2}{*}{Aleatoric Only (MCP)} & 92.6\% & 0.902 & 0.751 \\
 & {\scriptsize [91.0-94.0]} & {\scriptsize [0.875-0.926]} & {\scriptsize [0.713-0.789]} \\
\cmidrule{1-4}
\multirow{2}{*}{Epistemic Only (MCMD)} & 98.8\% & 0.831 & \textbf{0.788} \\
 & {\scriptsize [98.2-99.3]} & {\scriptsize [0.785-0.872]} & {\scriptsize [0.742-0.834]} \\
\cmidrule{1-4}
\multirow{2}{*}{Standard Uncertainty} & \textbf{99.3\%} & 0.824 & \textbf{0.788} \\
 & {\scriptsize [98.9-99.7]} & {\scriptsize [0.777-0.866]} & {\scriptsize [0.741-0.833]} \\
\midrule
\multirow{2}{*}{\textbf{Dual Veto (Hybrid)}} & 91.3\% & \textbf{0.913} & 0.755 \\
 & {\scriptsize [89.8-92.7]} & {\scriptsize [0.888-0.935]} & {\scriptsize [0.717-0.794]} \\
\bottomrule
\end{tabular}
}
\caption{Ablation of the deferral policy components on the MD-SN model ($\alpha=0.05$). 95\% CI calculated via empirical bootstrap resampling ($n=1,000$) are reported below each metric.
}
\label{tab:ablation_short}
\end{table}

\begin{table*}[ht!]
\centering
\resizebox{0.8\linewidth}{!}{
\small
\begin{tabular}{lcccccc}
\toprule
\textbf{Architecture} & \textbf{Risk Tol. ($\alpha$)} & \textbf{Coverage $\uparrow$} & \textbf{TPDR $\downarrow$} & \textbf{Clear $F_2 \uparrow$} & \textbf{Clear ECE $\downarrow$} & \textbf{Risk-Kappa $\uparrow$} \\
\midrule
\multirow{3}{*}{\textbf{Standard MIL}} 
& 0.01 & 63.5\% & 29.6\% & 0.979 & \textbf{0.009} & 0.546 \\
& 0.02 & 75.4\% & 19.4\% & 0.954 & 0.012 & 0.644 \\
& 0.05 & 91.1\% & \textbf{7.0\%} & 0.902 & 0.015 & 0.747 \\
\midrule
\multirow{3}{*}{\textbf{MD-SN MIL}} 
& 0.01 & 67.7\% & 33.2\% & \textbf{0.982} & 0.021 & 0.576 \\
& 0.02 & 77.4\% & 24.8\% & 0.969 & 0.028 & 0.665 \\
& 0.05 & \textbf{91.3\%} & 10.2\% & 0.913 & 0.029 & \textbf{0.755} \\
\bottomrule
\end{tabular}
}
\caption{Evaluation of the framework as an adjustable operational dial using the CV Ensemble UQ algorithm. By varying the conformal risk tolerance ($\alpha$), administrators can dynamically balance system Coverage against discriminative safety (Clear $F_2$) and the True Positive Deferral Rate (TPDR). \textbf{Note:} Full results, including 95\% bootstrap Confidence Intervals, are detailed in Appendix \ref{sec:results}.}
\label{tab:operational_dial}
\end{table*}

An ablation study (Table \ref{tab:ablation_short}) on the MD-SN architecture $(\alpha=0.05)$ confirms that single-modality uncertainty checks are structurally insufficient. Relying exclusively on aleatoric control (MCP) successfully flags probabilistic ambiguity but remains blind to overconfident geometric anomalies. Conversely, an epistemic-only policy (MCMD) suffers a catastrophic drop in safety by unsafely automating 98.8\% of the cohort. Crucially, this mirrors the failure of traditional selective classification approaches (represented here as ``Standard Uncertainty'' using a 90th-percentile predictive threshold), which yields a falsely optimistic 99.3\% coverage but degrades the Clear $F_2$ to 0.824.



Although 
single-modality policies yield nominally higher Custom Risk-Kappa scores (0.788), their 95\% CIs overlap extensively with our hybrid framework. 
Because of this overlap, the advantage of the dual veto on the Risk-Kappa metric is not statistically established at this current sample size. However, the hybrid policy achieves a statistically significant leap in Clear $F_2$ (0.913 vs. 0.824, with non-overlapping CIs). Therefore, the $\sim$8\% drop in coverage represents a necessary, statistically robust trade-off to eliminate the unsafe false negatives that standard UQ backends fail to intercept.


\subsection{Clinical Efficiency and the Risk Trade-off}

While strict safety constraints ($\alpha=0.01$) effectively isolate a highly reliable predictive domain, a static threshold fails to accommodate the fluctuating resource constraints of real-world healthcare systems. A clinically viable NLP triage framework must function as an adjustable operational dial, allowing hospital administrators to balance diagnostic safety with administrative workload reduction. Table \ref{tab:operational_dial} demonstrates the framework's scalability across broader risk thresholds ($\alpha \in \{0.01, 0.02, 0.05\}$).

Relaxing the conformal boundary on the MD-SN ensemble from a strict $\alpha=0.01$ to a moderate $\alpha=0.02$ allows the framework to expand its operational coverage to 77.4\%, while maintaining a robust Clear $F_2$ of 0.969. At a more lenient $\alpha=0.05$ setting, the system safely automates an outstanding 91.3\% of patient trajectories. Even at this high coverage level, the framework preserves a Clear $F_2$ of 0.913 and a low TPDR of 10.2\%.

Furthermore, evaluating the Standard MIL architecture reinforces the system's architecture-agnostic flexibility. At $\alpha=0.05$, the Standard MIL provides comparable coverage (91.1\%) with superior calibration (ECE: 0.015) and an exceptionally low TPDR (7.0\%), albeit with slightly lower discriminative safety. This performance parity proves that the dual-veto framework successfully rescues uncalibrated baselines and provides administrators with viable operational alternatives: prioritizing maximum safety (MD-SN MIL) or minimizing expert workload with superior calibration (Standard MIL).

\subsection{Interpretability of the Triage Decisions}
\label{sec:interpretabibity}

To visualize the underlying clinical reasoning and mitigate the black-box nature of deep neural networks, we analyzed feature 
attributions at the $\alpha=0.05$ threshold. Crucially, to accurately capture representational shifts, we explicitly compared the marginal feature attributions of the various test triage cohorts against the mean feature attributions of the validation cohort, which is our established, in-distribution baseline. While the absolute number of instances in these specific triage subsets is naturally small (e.g., 62 for automated FP, 17 for epistemic suspicion deferrals), they encompass the entirety of these edge cases within the test set\footnote{Aleatoric deferrals are excluded, since probabilistic ambiguity is determined by the joint conflict of typical characteristics, which marginal aggregations cannot adequately capture.}. A detailed methodological explanation of our custom hierarchical attribution system, alongside additional visual examples, is provided in Appendix~\ref{sec:shap}.

\begin{figure*}[t]
    \centering
    \begin{subfigure}[b]{0.48\textwidth}
        \centering
        \includegraphics[width=\textwidth]{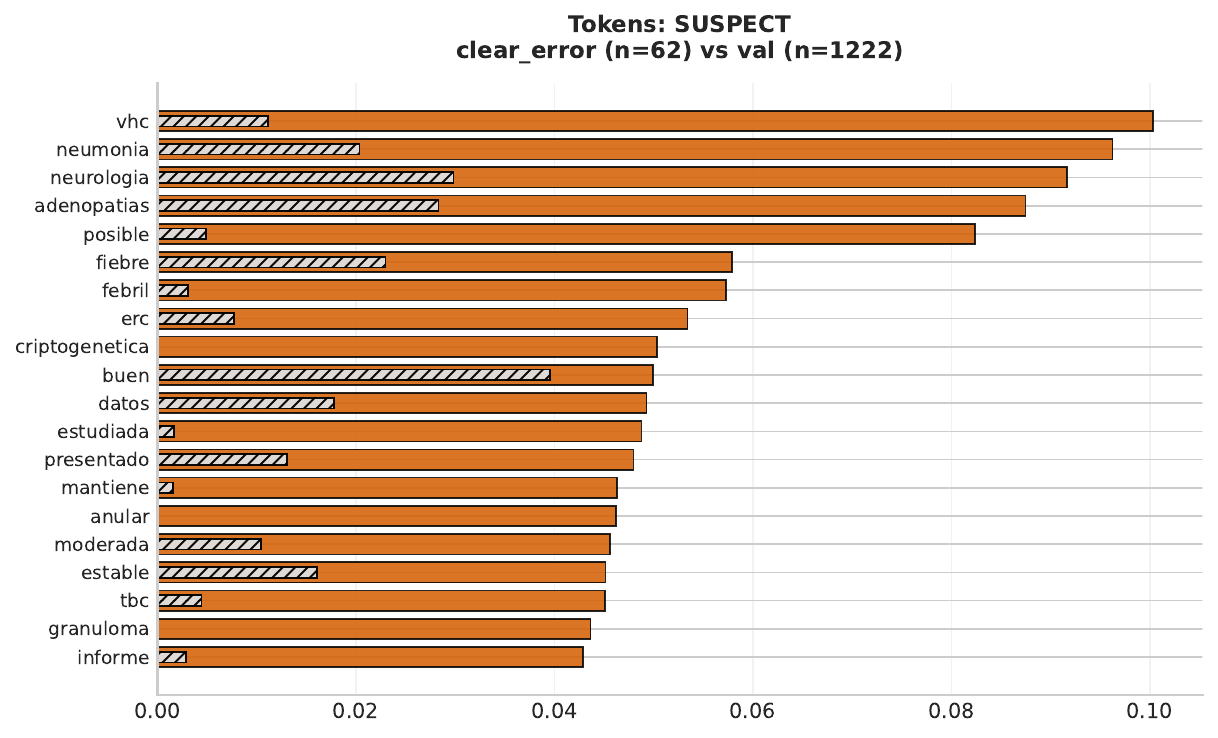} 
        \caption{Automated False Positives (``Clear Errors'')}
        \label{fig:shap_fp}
    \end{subfigure}
\hfill
    \begin{subfigure}[b]{0.48\textwidth}
        \centering
        \includegraphics[width=\textwidth]{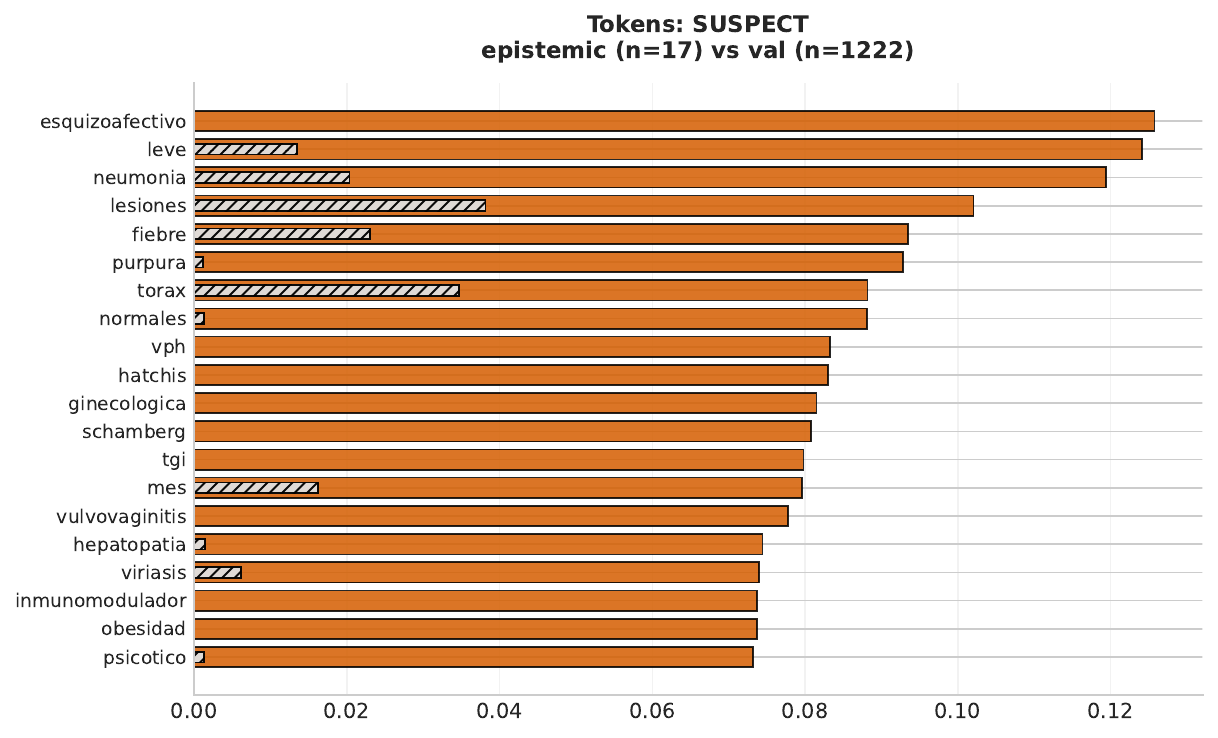} 
        \caption{Epistemic Deferrals (Suspicion)}
        \label{fig:shap_epistemic}
    \end{subfigure}
    
    \caption{Feature attribution comparison for complex triage edge cases at $\alpha=0.05$. Solid colored bars represent the mean token-level importance for the positive class on the designated test cohort. The underlying light grey bars with black diagonal hatching denote the validation baseline.}
    \label{fig:shap_main}
\end{figure*}

When analyzing the system's ``Clear Errors'', specifically FP where the model predicted clinical suspicion but no serology was documented, the feature attributions reveal strong clinical robustness when measured against this validation baseline. As shown in Figure \ref{fig:shap_fp}, these predictions are heavily driven by explicit indicator conditions mandating screening under EuroTEST guidelines, such as \textit{vhc} (Hepatitis C) and \textit{tbc} (Tuberculosis). This suggests the model successfully anchors on guideline-concordant phenotypic risk; the absence of a test highlights the noisy, partially observable nature of EHR documentation rather than a representational~failure.


Conversely, instances flagged by the epistemic geometric veto demonstrate severe representational collapse. As illustrated in Figure \ref{fig:shap_epistemic}, predictions for these deferred cases are hijacked by highly specific, OOD artifacts, such as psychiatric conditions (\textit{esquizoafectivo-schizoaffective-}, \textit{psicótico-psychotic-}) or unrelated localized infections (\textit{vulvovaginitis}), which registered near-zero predictive importance during validation. By calculating the Mahalanobis distance, the framework successfully recognizes this latent space distortion and safely vetoes the prediction before it can confidently misclassify the atypical presentation.



\section{Conclusion}
Translating biomedical NLP into clinical workflows is fundamentally limited by standard evaluation paradigms that force deterministic decisions on ambiguous narratives. This inherently produces overconfident misclassifications that directly conflict with medical safety requirements. To bridge this gap, we introduced and evaluated a risk-aware, hybrid selective screening framework for early HIV suspicion identification in a highly imbalanced Spanish EHR cohort.



Our empirical evaluations demonstrate that standard UQ methodologies conflate distinct predictive risks, forcing baseline models into severe coverage collapse when subjected to strict reliability constraints. In contrast, explicitly decoupling aleatoric uncertainty (via MCP) and epistemic uncertainty (via a MCMD veto) allows the evaluation framework to isolate a highly trustworthy operational domain. By functioning as an adjustable operational dial, the system enables healthcare administrators to dynamically balance diagnostic safety with administrative workload. Under strict safety constraints ($\alpha=0.01$), the dual-veto framework maximizes discriminative safety, achieving a high Clear $F_2$-score of 0.982 across 67.7\% of the cohort. When relaxed to accommodate high-throughput workloads ($\alpha=0.05$), the framework safely automates a 91.3\% of patient trajectories while preserving a robust 0.913 Clear $F_2$ and limiting true positive deferrals to just 10.2\%. Beyond quantitative safety metrics, our hierarchical feature attribution analysis confirms the clinical robustness of the triage logic: automated false positives reliably anchor on guideline-concordant phenotypic risks, while the epistemic veto proactively intercepts semantic artifacts and atypical presentations.
Ultimately, this study shows that optimizing biomedical NLP systems purely for raw classification accuracy creates a dangerous illusion of clinical readiness. By ensuring that highly ambiguous patient profiles and geometric anomalies are proactively flagged for expert review, our framework proves that rigorous, decoupled uncertainty quantification is a fundamental prerequisite for translating NLP into responsible clinical practice.

Building upon these findings, future work will focus on adapting this risk-aware triage paradigm to diverse clinical use cases where the asymmetric costs of errors differ from the preventative infectious disease model.

The complete UQ implementation pipeline and inference code is publicly available on GitHub\footnote{\href{https://github.com/romorale/mil_uq_public}{\texttt{https://github.com/romorale/mil\_uq\_public}}}.

\section*{Limitations}
While the framework bridges theoretical UQ and operational clinical safety, key methodological limitations remain. 
First, while the sensitivity analysis in Appendix \ref{sec:kappa_ablation} validates the robustness of the Custom Risk-Kappa under various simulated health-economic realities, it remains an experimental metric strictly tailored to the specific risk profile of early infectious disease screening. In this context, the severe epidemiological consequence of a missed diagnosis heavily outweighs the impact of an unnecessary serological test. However, applying this metric to domains with high physical intervention risks (e.g., invasive surgical biopsies in oncology) requires fundamental recalibration and formal health-economic validation.

Second, the static geometric thresholds $\tau_{dist}$ make the epistemic veto vulnerable to temporal or demographic dataset shifts prior to scheduled retraining. Third, although generative Large Language Models (LLMs) dominate current NLP benchmarks, extracting reliable, well-calibrated uncertainty from autoregressive generation remains an unsolved clinical challenge. Therefore, we explicitly prioritized a deterministic, spectrally normalized encoder to guarantee the geometric vetoes required for strict triage safety.

Finally, empirical validation is constrained to a single medical center. Given that HIV suspicion relies on partially observable behavioral indicators, cross-institutional validation is necessary to confirm the framework's broader generalizability.

\section*{Ethical Considerations}
The use of data for this study has been implemented in full compliance with strict adherence to the European General Data Protection Regulation (GDPR) and the Spanish Organic Law 3/2018 (LOPDGDD) on the Protection of Personal Data and Guarantee of Digital Rights. Due to strict institutional ethics board regulations privacy constraints regarding highly sensitive clinical narratives, 
the raw electronic health records 
cannot be made publicly accessible. 

\section*{Acknowledgements}

This work was supported by the projects EDHER-MED (PID2022-136522OB-C21; MCIN/AEI/10.13039/501100011-033/FEDER, UE), funded by the Spanish State Research Agency and the Spanish Ministry of Science, Innovation and Universities.



\bibliography{references}

\appendix
\counterwithin{figure}{section}
\counterwithin{table}{section}
\renewcommand\thefigure{\thesection\arabic{figure}}
\renewcommand\thetable{\thesection\arabic{table}}

\section{Hyperparameter Sensitivity Analysis}
\label{sec:hyperparameter_analysis}
To ensure the robustness of the epistemic veto, we evaluated the framework's triage efficiency across varying inertia convergence limits ($0.01$, $0.02$, $0.05$, $0.1$) and Mahalanobis distance percentiles ($90$th, $95$th, $99$th, $99.5$th, $99.9$th) on the test set, calibrated over the validation set. As shown in Table \ref{tab:hyperparameter_ablation}, the chosen configuration ($0.05$, $99$th) represents the empirical global optimum, maximizing the clinical safety metric (Custom Risk-Kappa) while maintaining a low TPDR.

\begin{table*}[t]
\centering
\resizebox{0.8\linewidth}{!}{
\small
\begin{tabular}{l c c c c c}
\toprule
\textbf{Inertia ($\Delta$)} & \textbf{Dist. Percentile} & \textbf{Coverage $\uparrow$} & \textbf{Clear $F_2 \uparrow$} & \textbf{TPDR $\downarrow$} & \textbf{Risk-Kappa $\uparrow$} \\
\midrule
\multicolumn{6}{l}{\textit{Strict Extreme}} \\
0.01 & 90th & 84.3\% & 0.946 & 19.7\% & 0.721 \\
\midrule
\multicolumn{6}{l}{\textit{Smooth Degradation}} \\
0.05 & 90th & 84.7\% & 0.945 & 19.7\% & 0.724 \\
0.05 & 95th & 87.5\% & 0.935 & 16.1\% & 0.746 \\
\midrule
\multicolumn{6}{l}{\textit{Chosen Optimum}} \\
\textbf{0.05} & \textbf{99th} & \textbf{91.3\%} & \textbf{0.913} & \textbf{10.1\%} & \textbf{0.755} \\
\midrule
\multicolumn{6}{l}{\textit{Loose Extreme}} \\
0.10 & 99.9th & 92.4\% & 0.904 & 8.7\% & 0.752 \\
\bottomrule
\end{tabular}
}
\caption{Hyperparameter Sensitivity Analysis across varying inertia convergence limits ($0.01$, $0.02$, $0.05$, $0.1$) and Mahalanobis distance percentiles ($90$th, $95$th, $99$th, $99.5$th, $99.9$th) on the test set. 
}
\label{tab:hyperparameter_ablation}
\end{table*}

\section{Custom Risk-Kappa ablation and the impact of model coverage}
\label{sec:kappa_ablation}

\begin{table*}[htbp]
\centering
\resizebox{1.0\linewidth}{!}{
\begin{tabular}{l | cccc | cc}
\toprule
\multirow{2}{*}{\textbf{Health-Economic Reality}} & \multicolumn{4}{c|}{\textbf{Penalty Weights}} & \textbf{Standard MIL} & \textbf{MDSN MIL} \\
 & \textbf{FN} & \textbf{FP} & \textbf{Def\_TN} & \textbf{Def\_TP} & \textbf{(63.5\% Cov.)} & \textbf{(67.7\% Cov.)} \\
\midrule
\textbf{Zero-Cost Deferral} \textit{(Maximum Caution)} & 1.0 & 1.0 & 0.0 & 0.0 & 0.948 & 0.960 \\
\textbf{Extreme Epidemiological} \textit{(Resource-Rich)} & 1.0 & 0.25 & 0.1 & 0.25 & 0.695 & 0.723 \\
\textbf{Default Baseline} \textit{(Preventative Medicine)} & \textbf{1.0} & \textbf{0.5} & \textbf{0.25} & \textbf{0.5} & \textbf{0.546} & \textbf{0.576} \\
\textbf{Symmetric Control} \textit{(Classification Baseline)} & 1.0 & 1.0 & 0.5 & 0.5 & 0.475 & 0.513 \\
\textbf{High Admin Burden} \textit{(Resource-Constrained)} & 1.0 & 0.75 & 0.5 & 0.75 & 0.432 & 0.463 \\
\bottomrule
\end{tabular}}
\caption{Custom Risk-Kappa Scores Across Simulated Health-Economic Realities.}
\label{tab:kappa_ablation}
\end{table*}

To evaluate the behavioral differences between the Standard MIL and the MD-SN MIL architectures under varying operational constraints, we conducted an ablation of the Custom Risk-Kappa metric across five simulated health-economic realities.

Table \ref{tab:kappa_ablation} demonstrates how these two architectures perform depending on the institutional penalties assigned to False Positives (FP), Deferred True Negatives (Def\_TN), and Deferred True Positives (Def\_TP). A missed diagnosis (False Negative) is consistently assigned the maximum catastrophic penalty of 1.0 across all scenarios.

Across all simulated environments, the MDSN MIL consistently outperforms the Standard MIL baseline. In a ``Zero-Cost Deferral'' scenario representing maximum caution (where manual hospital review incurs no penalty), the MDSN MIL achieves an exceptional Risk-Kappa of 0.960 compared to the Standard MIL's 0.948. This highlights that when an institution has unlimited bandwidth to review deferrals, the MDSN's rigid safety mechanisms provide near-perfect predictive accuracy on its covered subset.

Conversely, when simulating a resource-constrained environment (``High Administrative Burden''), penalties for false positives (0.75) and deferrals (0.5 to 0.75) are heavily increased. Even under these strict conditions, the MDSN MIL (0.463) maintains a clear advantage over the Standard MIL (0.432). This consistent superiority across both extreme epidemiological and highly constrained administrative realities proves that the MD-SN architecture's superior calibration and epistemic veto mechanism successfully isolate ambiguities without overwhelming institutional resources.

\section{Feature Attribution Analysis}
\label{sec:shap}

\subsection{Hierarchical Feature Attribution Methodology}
A common limitation in interpreting Multiple Instance Learning (MIL) architectures is the difficulty of projecting local token importances into the global document representation $(z_{doc})$. To generate the feature attributions, we implemented a custom hierarchical attention-occlusion explainer, computed in four stages:
\begin{enumerate}
    \item \textbf{Local Token Importance:} Token-level importance is extracted from the base transformer's last-layer attention weights (combining [CLS] and self-attention) and aggregated into continuous word-level scores.
    \item \textbf{Global MIL Aggregation:} To project local scores to the patient level, each word's score is multiplied by the normalized MIL gated-attention weight $(w_c)$ assigned to its respective chunk. The global importance is the sum of its MIL-weighted occurrences:
    \begin{equation*}
    I(\text{word}) = \sum_{c \in \text{chunks}} \text{Attn}(\text{word}_c) \cdot w_c
    \end{equation*}
    \item \textbf{Directional Impact (Occlusion):} Because attention only provides magnitude, we determine polarity via a leave-one-out occlusion test. Masking the target word across all chunks shifts the decision boundary; this logit margin $(\Delta)$ determines whether the token drives the prediction toward Suspicion or Non-Suspicion.
    \item \textbf{Cohort TF-IDF Aggregation:} To prevent highly-attended outlier documents from skewing the cohort average, patient-level importances are aggregated using a modified TF-IDF weighting. Term Frequency (intra-patient importance) is square-root scaled to dampen extremes, while the IDF penalty upweights consistently important terms across the cohort. Finally, these scores are averaged across the target subset to yield the stable attributions displayed in the figures.
\end{enumerate}

\begin{figure*}[t]
    \centering
    \begin{subfigure}[b]{0.48\textwidth}
        \centering
        \includegraphics[width=\textwidth]{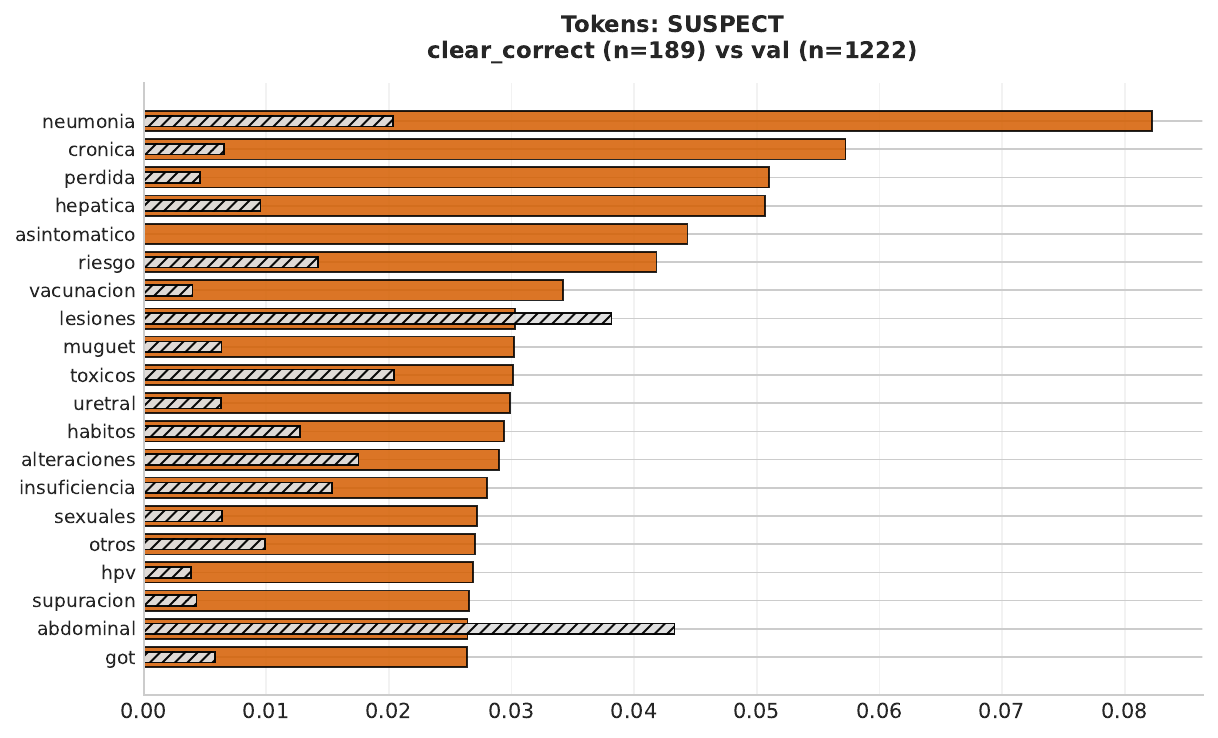} 
        \caption{Automated True Positives ("Clear Correct")}
        \label{fig:shap_tp}
    \end{subfigure}
    \hfill
    \begin{subfigure}[b]{0.48\textwidth}
        \centering
        \includegraphics[width=\textwidth]{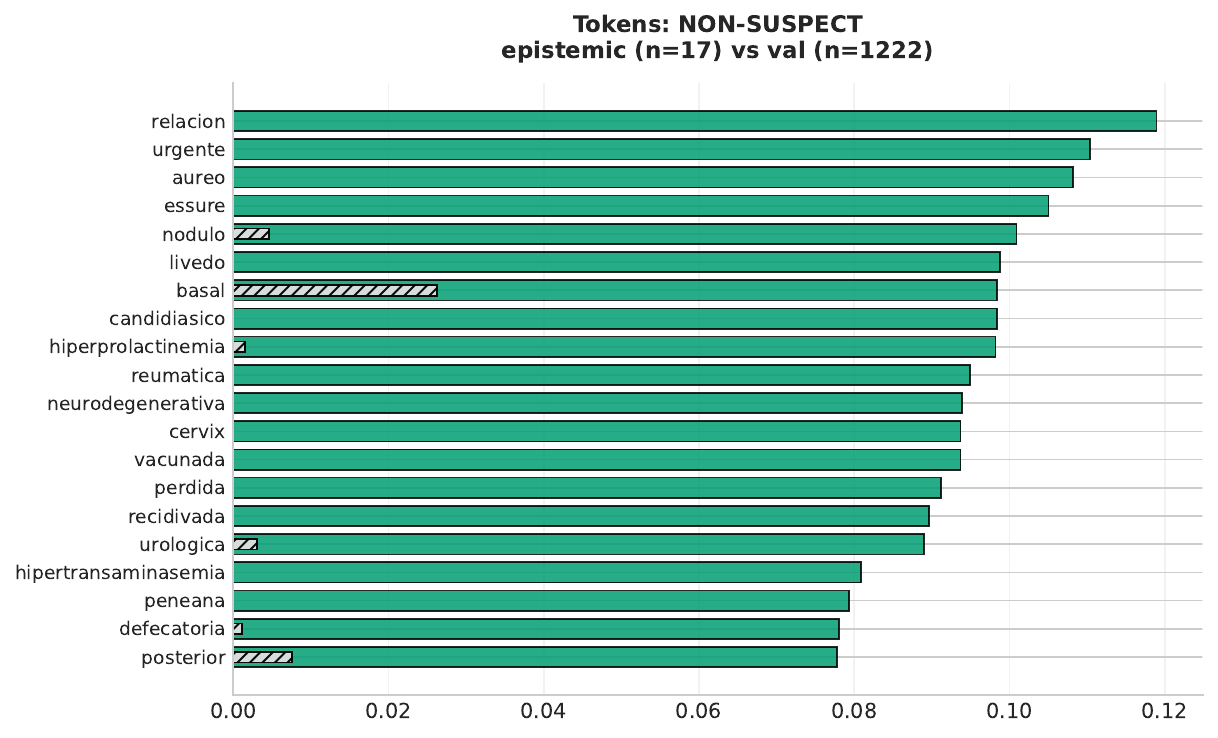} 
        \caption{Epistemic Deferrals (Non-Suspicion)}
        \label{fig:shap_epistemic_neg}
    \end{subfigure}
    
    \caption{Feature attribution comparison for complex triage edge cases at $\alpha=0.05$. Solid colored bars (orange for the positive class, green for the negative class) represent the mean token-level importance for the designated test cohort. The underlying light grey bars with black diagonal hatching denote the validation baseline.}
    \label{fig:shap_app}
\end{figure*}

\subsection{Supplementary Triage Visualizations}

This section provides supplementary visualizations to confirm the model's expected baseline behavior and further demonstrate the necessity of the epistemic veto.

Figure \ref{fig:shap_app} illustrates the feature attributions for the remaining triage cohorts. Plot \ref{fig:shap_tp} confirms the system's reliability on standard, safely automated true positive cases (``Clear Correct''). In these instances, the model's predictions rely entirely on established, in-distribution clinical indicators (such as \textit{neumonía}-\textit{pneumonia}- and \textit{muguet}), confirming that its automated decisions are grounded in relevant epidemiological phenotypes

Conversely, Plot \ref{fig:shap_epistemic_neg} demonstrates the critical role of the geometric veto even for cases where the model predicts the negative class (Non-Suspicion). In these deferred cases, the model is distracted by highly specific, OOD medical artifacts that distort the latent space. A prime example is the heavy attribution given to \textit{essure} (a specific gynecological implant). While such terms may spuriously co-occur with routine administrative testing in the training data, they are fundamentally unrelated to the targeted clinical construct. The MCMD successfully recognizes this geometric anomaly and intercepts the prediction, preventing the model from making an overconfident decision based on irrelevant clinical noise.

\section{Exhaustive Evaluation Metrics}
\label{sec:results}

\begin{table*}[ht!]
\centering
\resizebox{\textwidth}{!}{
\small
\setlength{\tabcolsep}{4pt}
\begin{tabular}{llcccccccc}
\toprule
\multirow{2}{*}{\textbf{Architecture}} & \multirow{2}{*}{\textbf{UQ Algorithm ($\alpha$)}} & \multicolumn{2}{c}{\textbf{Discriminative ($F_2\uparrow$)}} & \multicolumn{2}{c}{\textbf{Calibration (ECE$\downarrow$)}} & \multicolumn{4}{c}{\textbf{Triage Efficiency \& Safety}} \\
\cmidrule(lr){3-4} \cmidrule(lr){5-6} \cmidrule(lr){7-10}
 &  & \textbf{Binary} & \textbf{Clear} & \textbf{Binary} & \textbf{Clear} & \textbf{Coverage $\uparrow$} & \textbf{TPDR $\downarrow$} & \textbf{AURC $\downarrow$} & \textbf{Risk-Kappa $\uparrow$} \\
\midrule
\multirow{2}{*}{\textbf{Enc. Baseline}} & \multirow{2}{*}{CV Ens (0.01)} & 0.769 & 0.973 & 0.077 & 0.031 & 47.8\% & 50.5\% & 0.026 & 0.377 \\
 &  & {\scriptsize [0.722-0.811]} & {\scriptsize [0.958-0.985]} & {\scriptsize [0.063-0.092]} & {\scriptsize [0.021-0.042]} & {\scriptsize [45.2-50.3]} & {\scriptsize [43.9-57.3]} & {\scriptsize [0.020-0.032]} & {\scriptsize [0.341-0.410]} \\
\midrule
\multirow{8}{*}{\textbf{Standard MIL}} & \multirow{2}{*}{MC Dropout (0.01)} & 0.835 & 0.975 & 0.033 & 0.011 & 68.8\% & 34.2\% & 0.007 & 0.579 \\
 &  & {\scriptsize [0.793-0.874]} & {\scriptsize [0.954-0.991]} & {\scriptsize [0.024-0.043]} & {\scriptsize [0.006-0.016]} & {\scriptsize [66.4-71.2]} & {\scriptsize [28.2-40.5]} & {\scriptsize [0.005-0.009]} & {\scriptsize [0.543-0.616]} \\[1ex]
 & \multirow{2}{*}{CV Ens (0.01)} & 0.823 & 0.979 & \textbf{0.018} & \textbf{0.009} & 63.5\% & \textbf{29.6\%} & 0.007 & 0.546 \\
 &  & {\scriptsize [0.776-0.864]} & {\scriptsize [0.965-0.991]} & {\scriptsize [0.010-0.025]} & {\scriptsize [0.004-0.015]} & {\scriptsize [60.8-66.1]} & {\scriptsize [23.6-35.7]} & {\scriptsize [0.005-0.010]} & {\scriptsize [0.507-0.582]} \\
\cmidrule{2-10}
 & \multirow{2}{*}{CV Ens (0.02)} & 0.823 & 0.954 & 0.018 & 0.012 & 75.4\% & 19.4\% & 0.007 & 0.644 \\
 &  & {\scriptsize [0.776-0.864]} & {\scriptsize [0.931-0.973]} & {\scriptsize [0.010-0.025]} & {\scriptsize [0.006-0.019]} & {\scriptsize [73.0-77.6]} & {\scriptsize [14.4-24.6]} & {\scriptsize [0.005-0.010]} & {\scriptsize [0.606-0.679]} \\[1ex]
 & \multirow{2}{*}{CV Ens (0.05)} & 0.823 & 0.902 & 0.018 & 0.015 & 91.1\% & 7.0\% & 0.007 & 0.747 \\
 &  & {\scriptsize [0.776-0.864]} & {\scriptsize [0.872-0.929]} & {\scriptsize [0.010-0.025]} & {\scriptsize [0.008-0.024]} & {\scriptsize [89.7-92.6]} & {\scriptsize [3.9-10.5]} & {\scriptsize [0.005-0.010]} & {\scriptsize [0.705-0.783]} \\
\midrule
\multirow{10}{*}{\textbf{MD-SN MIL}} & \multirow{2}{*}{MC Dropout (0.01)} & \textbf{0.839} & 0.973 & 0.044 & 0.031 & 70.2\% & 30.6\% & 0.011 & \textbf{0.602} \\
 &  & {\scriptsize [0.797-0.877]} & {\scriptsize [0.955-0.988]} & {\scriptsize [0.034-0.054]} & {\scriptsize [0.024-0.039]} & {\scriptsize [67.8-72.5]} & {\scriptsize [24.7-37.0]} & {\scriptsize [0.006-0.016]} & {\scriptsize [0.566-0.639]} \\[1ex]
 & \multirow{2}{*}{MD-SN (0.01)} & 0.813 & 0.966 & 0.028 & 0.026 & \textbf{70.7\%} & 31.9\% & 0.012 & 0.601 \\
 &  & {\scriptsize [0.767-0.856]} & {\scriptsize [0.945-0.983]} & {\scriptsize [0.019-0.038]} & {\scriptsize [0.017-0.034]} & {\scriptsize [68.4-73.0]} & {\scriptsize [25.9-38.2]} & {\scriptsize [0.006-0.020]} & {\scriptsize [0.565-0.638]} \\[1ex]
 & \multirow{2}{*}{CV Ens (0.01)} & 0.821 & \textbf{0.982} & 0.022 & 0.021 & 67.7\% & 33.2\% & \textbf{0.006} & 0.576 \\
 &  & {\scriptsize [0.774-0.864]} & {\scriptsize [0.979-0.995]} & {\scriptsize [0.013-0.031]} & {\scriptsize [0.015-0.027]} & {\scriptsize [65.3-70.4]} & {\scriptsize [27.1-40.2]} & {\scriptsize [0.004-0.008]} & {\scriptsize [0.539-0.613]} \\
\cmidrule{2-10}
 & \multirow{2}{*}{CV Ens (0.02)} & 0.821 & 0.969 & 0.022 & 0.028 & 77.4\% & 24.8\% & 0.006 & 0.665 \\
 &  & {\scriptsize [0.774-0.864]} & {\scriptsize [0.954-0.983]} & {\scriptsize [0.013-0.031]} & {\scriptsize [0.021-0.036]} & {\scriptsize [75.2-79.8]} & {\scriptsize [19.4-30.7]} & {\scriptsize [0.004-0.008]} & {\scriptsize [0.629-0.701]} \\[1ex]
 & \multirow{2}{*}{CV Ens (0.05)} & 0.821 & 0.913 & 0.022 & 0.029 & 91.3\% & 10.2\% & 0.006 & 0.755 \\
 &  & {\scriptsize [0.774-0.864]} & {\scriptsize [0.888-0.935]} & {\scriptsize [0.013-0.031]} & {\scriptsize [0.021-0.038]} & {\scriptsize [89.8-92.7]} & {\scriptsize [6.2-14.2]} & {\scriptsize [0.004-0.008]} & {\scriptsize [0.717-0.794]} \\
\bottomrule
\end{tabular}
}
\caption{Discriminative performance, calibration, and triage efficiency across models and Uncertainty Quantification (UQ) backends. Metrics evaluate the forced deterministic output (Binary) and the automated subset after deferral (Clear). Triage safety is further assessed via Coverage, True Positive Deferral Rate (TPDR), Area Under the Risk-Coverage curve (AURC), and Custom Risk-Kappa. 95\% Confidence Intervals (CI) calculated via empirical bootstrap resampling ($n=1,000$) are reported below each metric. The best performance per metric in the strict safety tier ($\alpha=0.01$) is highlighted in bold.}
\label{tab:main_results}
\end{table*}

Table \ref{tab:main_results} provides the comprehensive triage metrics across all tested architectures, Uncertainty Quantification algorithms, and risk tolerances. All metrics include 95\% Confidence Intervals (CI) derived via empirical bootstrap resampling ($n=1,000$).

\end{document}